\documentclass[pdflatex,sn-mathphys-num]{sn-jnl}

\usepackage{graphicx}%
\usepackage{multirow}%
\usepackage{amsmath,amssymb,amsfonts}%
\usepackage{amsthm}%
\usepackage{mathrsfs}%
\usepackage[title]{appendix}%
\usepackage{xcolor}%
\usepackage{textcomp}%
\usepackage{manyfoot}%
\usepackage{booktabs}%
\usepackage{algorithm}%
\usepackage{algorithmicx}%
\usepackage{algpseudocode}%
\usepackage{listings}%
\usepackage{subcaption}

\theoremstyle{thmstyleone}%
\theoremstyle{thmstyletwo}%

\theoremstyle{thmstylethree}%

\usepackage{subcaption}
\begin{document}

\title[Article Title]{TopoCast: A Topological Fidelity Framework for Evaluating Transformer-Based Time Series Forecasting}


\author*[1]{\fnm{Sandeepa} \sur{Weerasekara}}\email{sandeepa.25@cse.mrt.ac.lk}

\author*[1]{\fnm{Sandareka} \sur{Wickramanayake}}\email{sandarekaw@cse.mrt.ac.lk }

\affil*[1]{\orgdiv{Department of Computer Science and Engineering}, \orgname{University of Moratuwa}, \orgaddress{\street{ Katubedda}, \city{Moratuwa}, \postcode{10400}, \country{Sri Lanka}}}


\abstract{

Deep learning-based models have achieved state-of-the-art performance in Time Series Forecasting (TSF), yet their evaluation remains dominated by pointwise error metrics such as Mean Squared Error (MSE), which quantify numerical accuracy but overlook structural properties of the forecast signal, including recurrent dynamics, oscillatory behavior, and phase alignment. As a result, forecasts exhibiting over-smoothing, phase shifts, or frequency distortions may achieve favorable error scores despite substantial structural degradation. To address this limitation, we propose \textbf{TopoCast}, a topology-driven framework for evaluating structural fidelity in TSF. TopoCast reconstructs phase-space representations of forecast and ground-truth sequences using Takens delay embedding and applies persistent homology to characterize their intrinsic dynamics. We derive four complementary topological fidelity measures from persistence diagrams and aggregate them into a \emph{Topological Fidelity Score (TFS)}. We further introduce \emph{dominant cycle overlap}, a novel metric that maps persistent topological features to the temporal domain to assess whether dominant oscillatory patterns occur at the correct time points. Combined with TFS, this yields the \emph{Localized Topological Fidelity Score (LTFS)}, a phase-aware measure that captures temporal localization errors invisible to existing evaluation metrics. Experiments on five Transformer architectures across three real-world benchmark datasets demonstrate that models with similar forecasting errors can exhibit markedly different structural fidelity profiles, revealing failure modes overlooked by conventional evaluation and highlighting the value of topology-aware forecast assessment.
}




\maketitle

\section{Introduction}
\label{sec: Intro}

Time series forecasting (TSF) is a fundamental task in many application domains, including energy systems~\cite{hossain2025time}, finance~\cite{praveen2025financial}, and healthcare~\cite{saleh2025multivariate}, where forecast quality directly influences downstream decision-making. Recent advances in deep learning have established Transformer-based architectures as leading approaches for TSF due to their ability to model long-range temporal dependencies and complex sequential patterns~\cite{wu2021autoformer,zhou2022fedformer,zhou2021informer,nie2022time}. Despite their strong predictive performance, model evaluation remains largely dependent on pointwise error metrics such as Mean Squared Error (MSE) and Mean Absolute Error (MAE).

Although widely adopted, pointwise metrics assess only numerical discrepancies between predicted and observed values at individual time steps. Consequently, they provide limited insight into whether a forecast preserves the underlying temporal structure of the target signal. Forecasts that closely match the overall trend may nevertheless distort seasonal patterns, introduce temporal lag, suppress high-frequency dynamics, or generate spurious oscillatory behaviour. Such deviations alter important characteristics of the underlying dynamical system, yet often result in only minor changes in pointwise error measures~\cite{yu2025towards}. As a result, structurally distinct forecasts may receive similar evaluation scores despite exhibiting substantially different temporal behaviours.

\begin{figure}[t]
    \centering

    \begin{subfigure}[b]{0.49\linewidth}
        \centering
        \includegraphics[width=\linewidth]{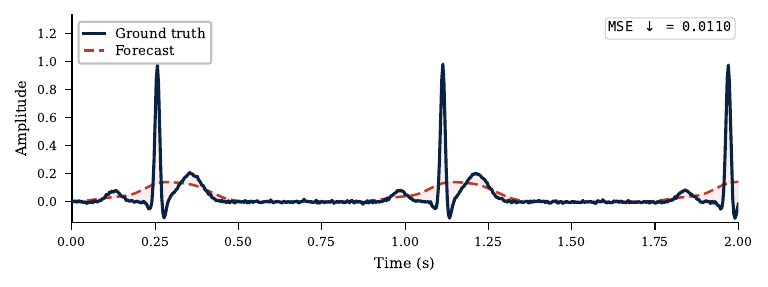}
        \caption*{(A) Smooth forecast}
    \end{subfigure}
    \hfill
    \begin{subfigure}[b]{0.49\linewidth}
        \centering
        \includegraphics[width=\linewidth]{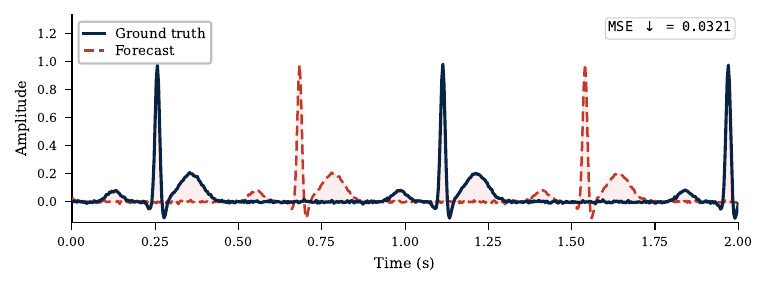}
        \caption*{(B) Phase shift}
    \end{subfigure}

    \caption{Two structural failure modes undetectable by MSE. Both forecasts would be considered acceptable under pointwise evaluation, yet they exhibit fundamentally different and complementary forms of structural degradation.}
    \label{fig:motivation}
\end{figure}

Figure~\ref{fig:motivation} illustrates two representative examples. In Figure~\ref{fig:motivation}(A), a smoothed forecast achieves low MSE while failing to preserve the sharp R-peaks that characterize the ECG signal. In Figure~\ref{fig:motivation}(B), the forecast reproduces the dominant oscillatory pattern with comparable amplitude and frequency but exhibits a temporal phase shift relative to the ground truth. Although these forecasts differ substantially in their structural properties, both may be assigned favourable scores under conventional pointwise evaluation. These observations motivate the need for evaluation methodologies that explicitly quantify structural preservation in forecasted time series.

Topological Data Analysis (TDA), particularly persistent homology, provides a principled framework for characterizing the geometric and dynamical structure of time series. By reconstructing the underlying phase space through Takens delay embedding~\cite{takens2006detecting} and tracking the emergence and disappearance of topological features across multiple scales, persistent homology captures recurrent structures, cycles, and other dynamical properties that are not reflected in pointwise error metrics~\cite{lin2025time}. The resulting topological descriptors are robust to small perturbations and provide a complementary perspective on forecast quality.

In this work, we propose \textbf{TopoCast}, a persistent homology-based framework for evaluating structural fidelity in time series forecasting. Given a forecast sequence and its corresponding ground truth, we construct multivariate Takens embeddings and extract persistence diagrams. From these diagrams, we derive four complementary fidelity components, which quantify loop count, dominant cycle strength, total persistence, and diagram complexity, respectively. These components are aggregated using the geometric mean to obtain a \emph{Topological Fidelity Score} (TFS). To assess temporal localization of recurrent structures, we further introduce \emph{dominant cycle overlap}, a novel metric that maps cocycle generators back to the original temporal domain through inverse Takens reconstruction. This metric quantifies the extent to which dominant oscillatory structures in the forecast align with those in the ground truth. By integrating dominant cycle overlap with TFS, we obtain the \emph{Localized Topological Fidelity Score} (LTFS), a phase-aware measure that captures temporal localization errors not reflected by existing diagram-level metrics, including persistence summaries and Wasserstein distances.

We evaluate TopoCast on five Transformer-based forecasting models, namely Autoformer~\cite{wu2021autoformer}, FEDformer~\cite{zhou2022fedformer}, Informer~\cite{zhou2021informer}, Transformer~\cite{vaswani2017attention}, and PatchTST~\cite{nie2022time}, using three real-world multivariate benchmark datasets (ETTm2, Exchange Rate, and ILI) across multiple forecasting horizons. Experimental results demonstrate that models with comparable forecasting errors can exhibit substantially different topological fidelity profiles. Furthermore, LTFS reveals temporal phase-related failure modes that remain undetected by both pointwise and existing diagram-level evaluation methods.

\vspace{0.5em}
\noindent The main contributions of this work are summarized as follows:

\begin{enumerate}
\item We propose \textbf{TopoCast}, a persistent homology-based framework for evaluating structural fidelity in time series forecasting, providing information complementary to conventional pointwise error metrics while requiring access only to forecast outputs.
\item We introduce \emph{dominant cycle overlap}, a novel topological metric that quantifies temporal localization of dominant oscillatory structures by mapping persistent homology generators back to the time domain, enabling the detection of phase-related forecasting errors that are not captured by existing diagram-level evaluation methods.
\end{enumerate}

\section{Related Work}\label{sec1}

\subsection{Transformer-Based Time Series Forecasting}
Transformer architectures~\cite{vaswani2017attention} have become a dominant paradigm for time series forecasting owing to their capacity to model long-range temporal dependencies via self-attention. Several variants address the quadratic complexity 
of standard attention~\cite{child2019generating, 
zhou2021informer, kitaev2020reformer}, while others explicitly 
target temporal structure: Autoformer~\cite{wu2021autoformer} 
replaces dot-product attention with an autocorrelation mechanism coupled with trend--seasonal decomposition, 
FEDformer~\cite{zhou2022fedformer} leverages Fourier and wavelet transforms for frequency-domain modelling, and PatchTST~\cite{nie2022time} introduced patch-based tokenisation to improve local semantic representation. Despite their strong predictive performance, these models are evaluated almost exclusively on pointwise error, and whether their forecasts preserve the structural and dynamical properties of the target 
signal remains an open and largely unexamined question.

\subsection{Limitations of Forecasting Evaluation Metrics}
Pointwise metrics such as MSE and MAE remain the dominant evaluation criteria in time series 
forecasting~\cite{hyndman2006another, makridakis2020m4, 
wen2022transformers}, yet their limitations as sole measures are increasingly recognised. Both reduce forecast quality to an average of per-timestep deviations, rendering them blind to structurally meaningful discrepancies: a forecast may align 
numerically with the ground truth while failing to reproduce its 
directionality, variance, or periodic pattern. Patch-wise 
structural loss formulations~\cite{kudrat2025patch} have 
highlighted that MSE's point-independence assumption causes it to 
overlook correlation structure, variance dynamics, and phase 
relationships simultaneously, and a recent re-examination of 
long-term forecasting evaluation~\cite{phungtua2026we} showed 
that aggregated pointwise metrics obscure window-level 
heterogeneity, misrepresenting model behaviour on structurally 
irregular segments.

To address these limitations, shape-aware and alignment metrics 
have been proposed. Dynamic Time Warping and 
Soft-DTW~\cite{sakoe1978dynamic, cuturi2017soft} compare time 
series by allowing temporal distortions, the Temporal Distortion 
Index~\cite{le2019shape} quantifies timing discrepancies based on 
the DTW warping path, and Wasserstein distance evaluates 
distributional similarity between forecast and observed series. 
Yu et al.~\cite{yu2025towards} demonstrated that two time series 
with identical MSE can have entirely different geometric 
structures, directly motivating structure-aware evaluation. 
Despite capturing complementary aspects of forecast quality, 
none of these metrics evaluates whether a forecast preserves the 
underlying topological structure of the signal — its recurrent 
cycles, dominant seasonal patterns, and broader geometric 
characteristics — motivating a topology-aware evaluation 
framework grounded in persistent homology.

\subsection{Topological Data Analysis for Time Series}
Topological Data Analysis (TDA)~\cite{carlsson2009topology} 
provides a mathematically principled framework for characterising the global, scale-invariant structure of data. Persistent homology (PH)~\cite{edelsbrunner2002topological}, its core 
computational tool, tracks the birth and death of topological features — connected components, loops, and voids — as a filtration parameter varies, producing a persistence diagram encoding the multi-scale shape of the data. The Stability 
Theorem~\cite{cohen2005stability} guarantees that small perturbations produce bounded changes in the persistence diagram, making PH inherently robust to noise. Efficient implementations 
including Ripser~\cite{bauer2021ripser}, GUDHI, and 
Giotto-TDA~\cite{tauzin2021giotto} have made large-scale computation practical, and TDA has demonstrated the ability to detect structural patterns invisible to conventional statistical descriptors across domains, including materials science~\cite{obayashi2022homcloud} and financial 
modelling~\cite{gidea2018topological}.

In time series analysis, PH is typically applied via 
delay-coordinate embeddings that lift signals into point clouds for topological feature extraction~\cite{perea2015sliding}. The 
Takens embedding theorem~\cite{takens1981detecting} guarantees that such embeddings are diffeomorphic to the underlying attractor, preserving its topological invariants. In deep learning, PH has served as a feature extractor for classification~\cite{gidea2018topological}, a contrastive learning signal~\cite{kim2026topocl}, and a basis for graph ensembles in regression tasks~\cite{nguyen2025persistent}. 
Domain-adaptive extensions~\cite{heo2024persistent} and supervised forecasting frameworks~\cite{lin2025time} have further 
demonstrated topology's utility as a predictive signal, though the latter treats topology as a learning feature rather than an evaluation criterion, leaving structural fidelity and temporal localisation largely unaddressed.

\section{TopoCast Methodology}
\label{sec3}

\begin{figure}[t]
    \centering
    \includegraphics[width=0.95\linewidth]{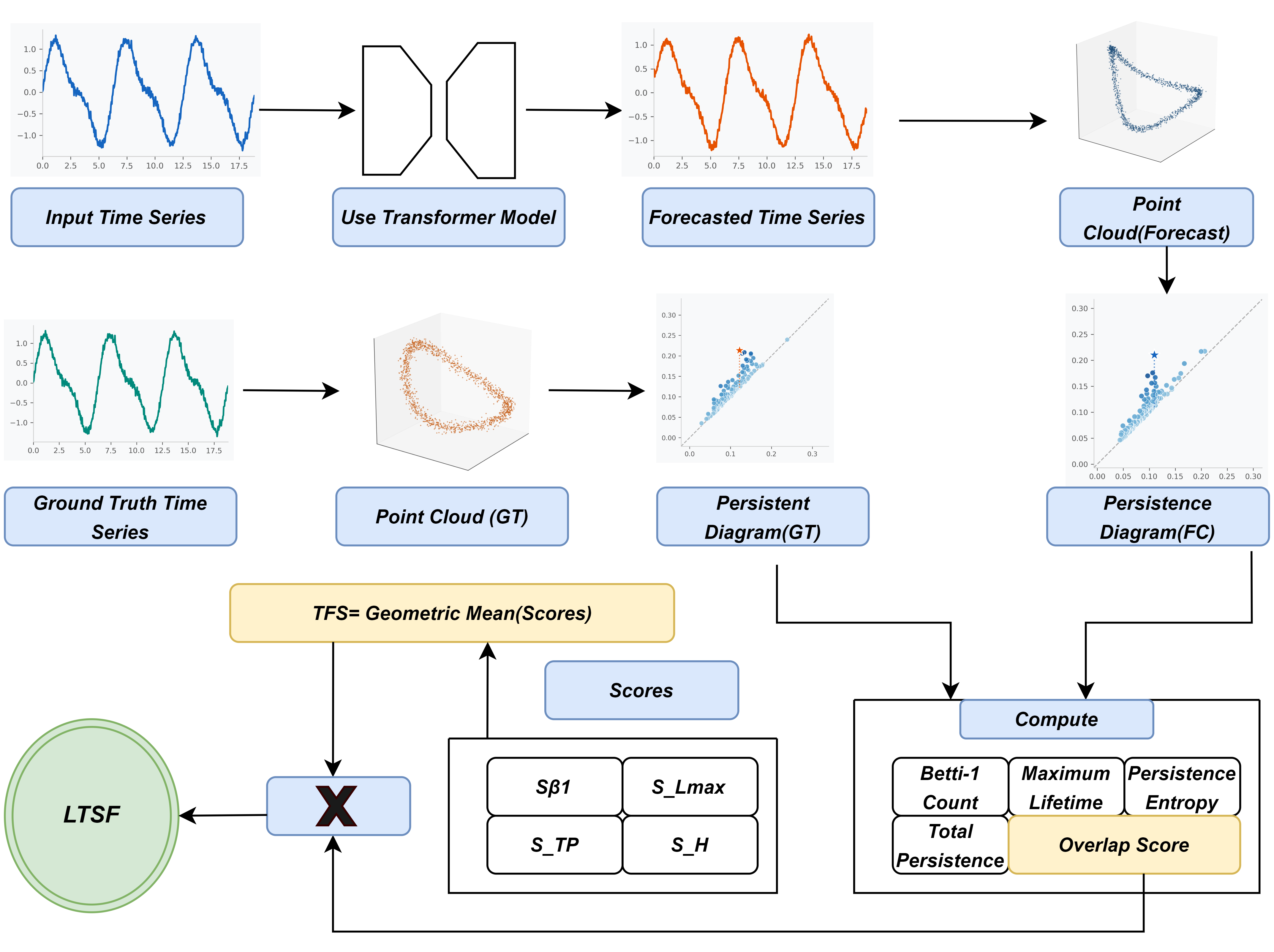}
    \caption{The TopoCast structural fidelity evaluation pipeline.}
    \label{fig:topocast_pipeline}
\end{figure}

Figure~\ref{fig:topocast_pipeline} illustrates the complete TopoCast evaluation pipeline described in the following subsections.

\subsection{Phase Space Reconstruction}
\label{sec:phase_space}

A time series cannot be directly analysed topologically in its raw form because one-dimensional signals have trivial topology. 
The key step is to lift the signal into a higher-dimensional 
phase space where its dynamical structure, including cycles, 
attractors, and recurrent patterns, becomes geometrically visible 
as loops and connected components.

The Takens embedding theorem~\cite{takens1981detecting} guarantees 
that a delay-coordinate embedding reconstructs the topology of 
the underlying dynamical attractor. For a multivariate signal 
with $C$ channels, each channel $x_i(t)$ is embedded 
independently with dimension $m = 3$ and delay $\tau = 2$, 
producing an embedding matrix for each channel:
\begin{equation}
    \mathbf{X}_i = \bigl[x_i(t),\; x_i(t+\tau),\; x_i(t+2\tau)\bigr],
    \quad t = 0, 1, \ldots, N-1
    \label{eq:takens}
\end{equation}
where $N = T - (m-1)\tau = T - 4$ and $T$ is the sequence 
length. The $C$ channel embeddings are concatenated column-wise 
to produce a single joint point cloud $\mathcal{X} \in 
\mathbb{R}^{N \times mC}$. Both the forecast $\hat{y}$ and 
ground truth $y$ are embedded independently using identical 
parameters, producing point clouds $\mathcal{X}^{\hat{y}}$ and 
$\mathcal{X}^{y}$ of the same dimension.

\begin{figure}[h]
    \centering
    \includegraphics[width=\linewidth]{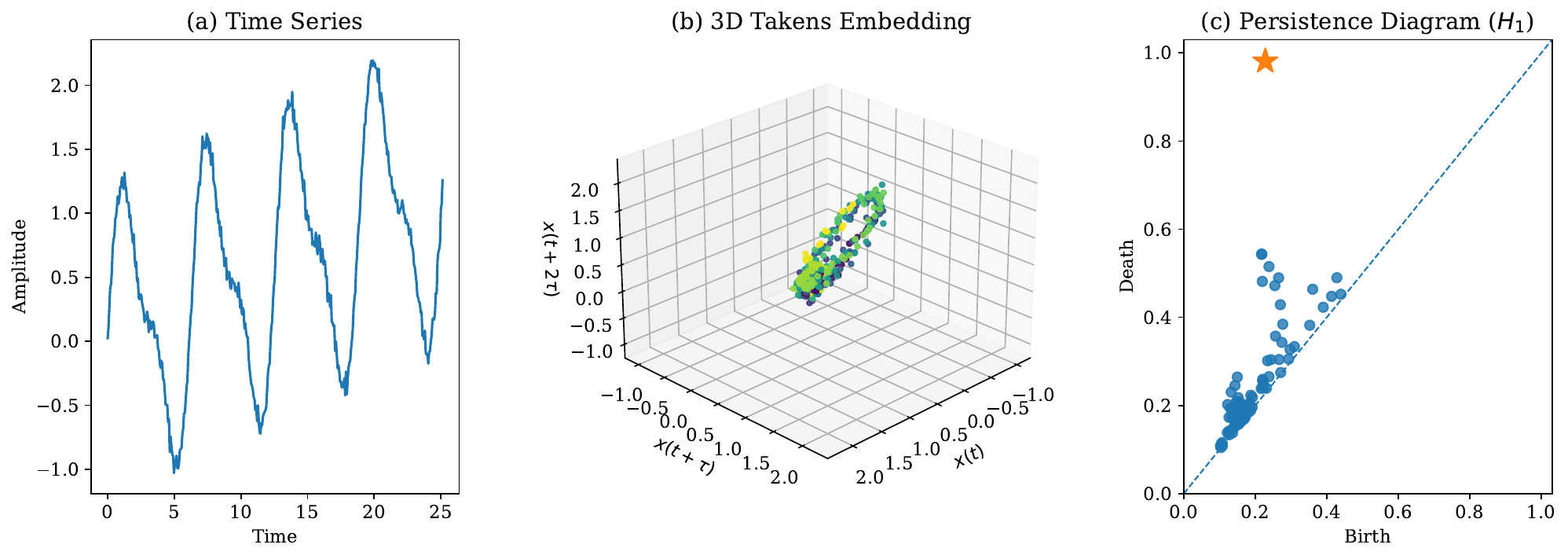}
    \caption{From time series to topological descriptor via Takens delay embedding and $H_1$ persistent homology.}
    \label{fig:tda_pipeline}
\end{figure}

The embedding dimension $m = 3$ and delay $\tau = 2$ are fixed 
throughout all experiments. These values are standard choices for 
short-to-medium length financial and energy time series and are 
consistent with prior work applying persistent homology to time 
series forecasting
evaluation~\cite{perea2015sliding,myers2023persistence}. 
Figure~\ref{fig:tda_pipeline} illustrates this pipeline for a 
synthetic cyclic signal, showing how a periodic time series 
forms a closed loop in phase space whose dominant feature is 
captured as a single high-persistence point in the $H_1$ diagram.

\subsection{Persistent Homology Extraction}
\label{sec:ph_extraction}

A Vietoris--Rips filtration is constructed on the pairwise 
Euclidean distance matrix of each point cloud. The filtration works by progressively connecting points as the distance threshold $\varepsilon$ increases: at each value of $\varepsilon$, 
any two points within distance $\varepsilon$ are connected by an 
edge, and any set of points mutually within distance $\varepsilon$ 
forms a simplex. As $\varepsilon$ grows from $0$ to $\infty$, 
topological features including connected components and loops are 
born and die.

$H_1$ persistence features, which are one-dimensional loops, are 
the primary descriptor because periodic signals form closed loop 
structures under Takens embedding. A time series with a dominant 
seasonal cycle of period $P$ will trace a closed loop in phase 
space, whose lifetime in the persistence diagram reflects the 
strength and regularity of that cycle. Each $H_1$ feature is 
tracked as a birth--death interval $[b, d)$ with lifetime 
$\ell = d - b$. Features with $\ell < \varepsilon_0 = 10^{-6}$ 
are discarded as numerical noise.

Persistent homology is computed via the Ripser 
library~\cite{bauer2021ripser}, enabling both the persistence 
diagram and the representative cocycle of each feature to be 
extracted. The cocycle is required for dominant cycle 
localisation described in Section~\ref{sec:overlap}.

From each $H_1$ diagram, four scalar descriptors are computed:

\begin{itemize}
    \item $\beta_1$ \textbf{(loop count)}: the total number of 
    significant $H_1$ features, measuring how many independent 
    cyclic structures the signal contains.

    \item $L_{\max}$ \textbf{(dominant cycle strength)}: the 
    maximum lifetime $\max_k \ell_k$, measuring the strength of 
    the most persistent seasonal pattern.

    \item $TP$ \textbf{(total persistence)}: the sum of all 
    lifetimes $\sum_k \ell_k$, measuring the aggregate 
    topological energy across all cyclic structures.

    \item $H$ \textbf{(persistence entropy)}: $-\sum_k p_k 
    \log p_k$ where $p_k = \ell_k / \sum_j \ell_j$, measuring 
    the complexity of the loop lifetime distribution.
\end{itemize}

\noindent These four descriptors are computed independently for 
both $\mathcal{X}^{\hat{y}}$ and $\mathcal{X}^{y}$, yielding 
descriptor vectors for the forecast and ground truth respectively.

\subsection{Topological Fidelity Scores}
\label{sec:scores}

Four interpretable scores compare forecast descriptors against 
ground-truth descriptors, each targeting a distinct structural 
property (Table~\ref{tab:topo_scores}). All four use a log-ratio 
form bounded in $(0, 1]$, with $1.0$ indicating perfect 
preservation of the corresponding descriptor. They are combined 
via geometric mean into a composite \emph{Topological Fidelity Score}:
\begin{equation}
    TFS =
    \left(
        S_{\beta_1}
        \cdot
        S_{L_{\max}}
        \cdot
        S_{TP}
        \cdot
        S_H
    \right)^{1/4}
    \label{eq:stfs}
\end{equation}
The geometric mean ensures that collapse in any single component 
propagates aggressively to the composite, producing an 
unambiguous structural damage signal. TFS is 
further multiplied by the dominant cycle overlap 
(Section~\ref{sec:overlap}) to produce the final metric:
\begin{equation}
    LTFS
    =
    S_{\mathrm{TFS}}
    \times
    \text{Overlap}
    \label{eq:sltfs}
\end{equation}
$S_{\mathrm{LTFS}} \in [0,1]$, with $1.0$ indicating complete 
structural preservation and perfect temporal co-localisation of 
the dominant cycle. All scores are computed per test window and 
averaged over $N$ windows.

\begin{table}[ht]
\caption{Structural quality score components.}
\label{tab:topo_scores}
\begin{tabular}{@{}llp{5.5cm}@{}}
\toprule
Score & Formula & Interpretation \\
\midrule
$S_{\beta_1}$ &
    $\exp\!\left(-\left|\log\dfrac{\hat{\beta}_1}
                                  {\beta_1^{y}}\right|\right)$ &
    Loop count preservation \\[10pt]

$S_{L_{\max}}$ &
    $\exp\!\left(-\left|\log\dfrac{\hat{L}_{\max}}
                                  {L_{\max}^{y}}\right|\right)$ &
    Dominant cycle strength preservation \\[10pt]

$S_{TP}$ &
    $\exp\!\left(-\left|\log\dfrac{\widehat{TP}}
                                  {TP^{y}}\right|\right)$ &
    Total topological energy preservation \\[10pt]

$S_H$ &
    $\exp\!\left(-\left|\log\dfrac{\hat{H}}
                                  {H^{y}}\right|\right)$ &
    Persistence diagram complexity preservation \\
\botrule
\end{tabular}
\end{table}

\subsection{Dominant Cycle Overlap}
\label{sec:overlap}

TFS operates on diagram-level summaries and is 
therefore blind to temporal localisation errors: a forecast may 
achieve high TFS while systematically misplacing 
the dominant oscillation in time. Dominant cycle overlap addresses 
this by mapping the topological generator of the dominant $H_1$ 
feature back to the original time domain.

For a point cloud $\mathcal{X} \in \mathbb{R}^{N \times mC}$ 
constructed via Takens embedding, each row $i$ corresponds to 
time step $i$ in the original signal. The dominant cycle is 
identified as the longest-lived $H_1$ feature; its representative 
cocycle, extracted via Ripser, is a set of simplices whose vertex 
indices map injectively back to time steps in 
$\{0, 1, \ldots, T-1\}$. Applying this inverse mapping to both 
point clouds yields two sets of active time steps:
\begin{equation}
    \mathcal{T}^{y}_{\text{dom}} \subset \{0, 1, \ldots, T-1\},
    \qquad
    \mathcal{T}^{\hat{y}}_{\text{dom}} \subset \{0, 1, \ldots, T-1\}
\end{equation}
Temporal co-localisation is then quantified as the Jaccard 
coefficient between these two sets:
\begin{equation}
    \text{Overlap} =
    \frac{
        \left|
        \mathcal{T}^{\hat{y}}_{\text{dom}}
        \cap
        \mathcal{T}^{y}_{\text{dom}}
        \right|
    }{
        \left|
        \mathcal{T}^{\hat{y}}_{\text{dom}}
        \cup
        \mathcal{T}^{y}_{\text{dom}}
        \right|
    }
    \label{eq:jaccard}
\end{equation}
$\text{Overlap} \in [0,1]$, with $1.0$ indicating perfect 
co-localisation and $0.0$ indicating complete displacement. 
Windows where the ground-truth diagram contains no valid $H_1$ 
generator are excluded; windows where the forecast contains none 
are assigned $\text{Overlap} = 0$, representing complete dominant 
cycle loss. Dominant cycle overlap is the only TopoCast component 
that reconnects topological analysis to the original time domain, 
detecting phase localisation errors entirely invisible to 
TFS, the Wasserstein distance $W_2$, and all 
other diagram-level summaries.

%
%


\section{Experiments}
\label{sec:experiments}
\subsection{Datasets}
\label{subsec:datasets}
We evaluate TopoCast on three multivariate time series benchmarks spanning
diverse temporal dynamics and topological regimes. \textbf{ETTm2} consists
of 7 variables sampled at 15-minute intervals over two years, exhibiting
strong periodicity and long-range seasonal dependencies. \textbf{ILI}
(Influenza-Like Illness) is a weekly CDC dataset covering 2002--2021 with
7 variables and 966 time steps, characterised by higher volatility and
weaker seasonality than ETTm2. \textbf{Exchange} contains daily exchange
rates for eight currencies against the U.S. dollar, with non-stationary
dynamics and minimal periodic structure.

\subsection{Models}
\label{subsec:models}
We evaluate five Transformer-based forecasting architectures spanning
three categories of inductive bias. \textbf{Transformer}~\cite{vaswani2017attention}
serves as the standard self-attention baseline.
\textbf{Informer}~\cite{zhou2021informer} introduces ProbSparse attention
with encoder distillation for efficient long-horizon forecasting.
\textbf{Autoformer}~\cite{wu2021autoformer} replaces dot-product attention
with an autocorrelation mechanism coupled with trend--seasonal decomposition.
\textbf{FEDformer}~\cite{zhou2022fedformer} extends decomposition-based
modelling to the frequency domain via Fourier and wavelet transforms.
\textbf{PatchTST}~\cite{nie2022time} tokenises time series patches and
applies channel-independent self-attention across patches rather than
timesteps. Together they represent standard attention (Transformer, Informer),
decomposition-based attention (Autoformer, FEDformer), and patch-based
channel-independent attention (PatchTST).

\subsection{Implementation Details}
\label{subsec:implementation}
All experiments use official or widely adopted open-source 
PyTorch implementations trained under their original 
hyperparameter settings. Takens embedding uses $m = 3$, 
$\tau = 2$ throughout; persistent homology is computed via 
Ripser~\cite{bauer2021ripser} with cocycle extraction enabled. 
We evaluate 100 randomly sampled non-overlapping test windows 
per (model, dataset, prediction length) for ETTm2 and Exchange, 
and 50 for ILI given its smaller size. Prediction lengths are 
$\{48, 96, 192\}$ for ETTm2 and Exchange and $\{36, 48, 60\}$ 
for ILI, yielding 45 experiments (5 models $\times$ 3 datasets 
$\times$ 3 prediction lengths).

\section{Synthetic Validation}
\label{sec:synthetic}
Benchmark evaluations cannot verify that topological metrics 
respond correctly to specific failure modes, since real datasets 
provide no ground truth about which structural property has 
degraded. We therefore design a controlled synthetic validation 
where each failure mode is specified exactly by construction.

\subsection{Experimental Setup and Results}
\label{sec:synthetic_results}
\begin{figure}[htbp]
    \centering
    \includegraphics[width=\linewidth]{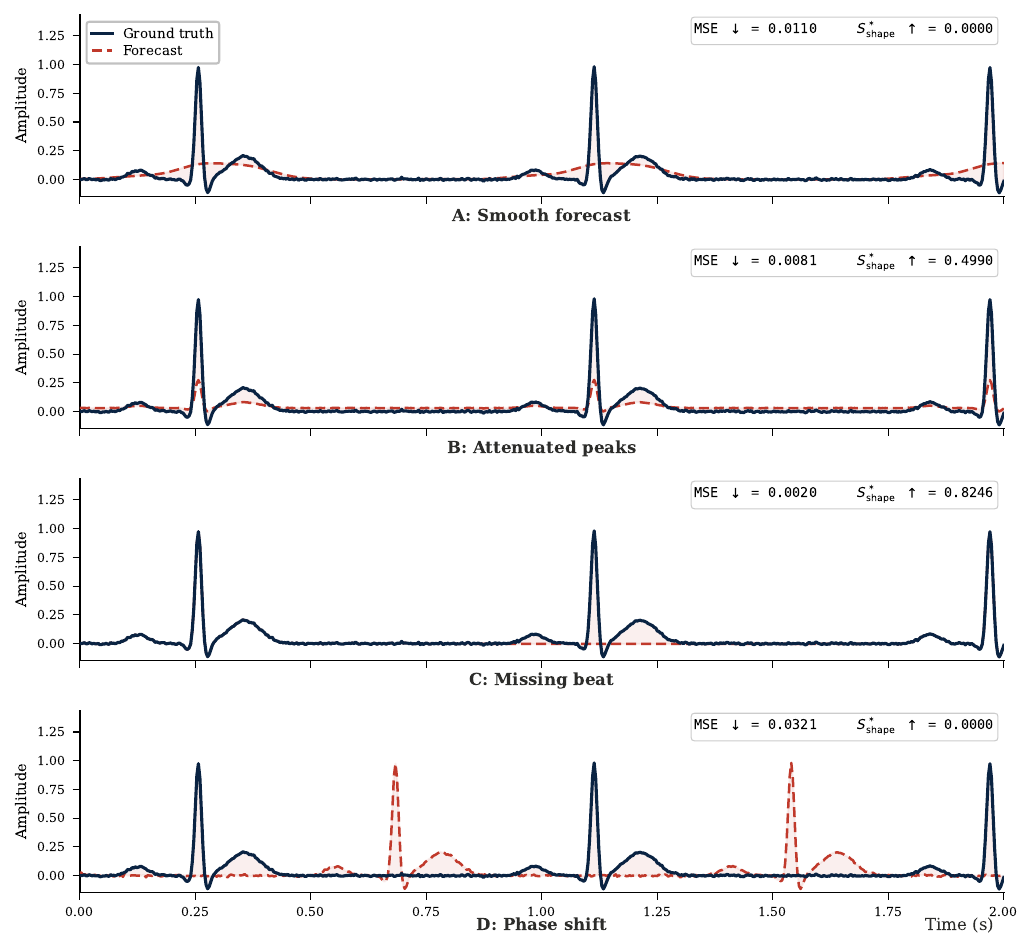}
    \caption{Synthetic ECG validation across four structural 
    failure scenarios. MSE remains low in all cases while LTFS correctly reflects the degree of structural  degradation in each.}
    \label{fig:ecg_synthetic}
\end{figure}

The ground truth is a synthetic ECG signal constructed via a 
sum-of-Gaussians PQRST template with 8 beats at $f_s = 300$\,Hz 
and heart rate of 70\,bpm, with low-amplitude Gaussian noise 
($\sigma = 0.005$) added to simulate measurement variability. 
Four forecast scenarios isolate distinct structural failure modes 
while keeping MSE deliberately low (Table~\ref{tab:ecg_scenarios}); 
the scenarios are visualised in Figure~\ref{fig:ecg_synthetic}.

\begin{table}[!t]
\caption{Controlled ECG forecast scenarios and targeted 
failure modes.}
\label{tab:ecg_scenarios}
\begin{tabular}{@{}lp{4.5cm}p{5cm}@{}}
\toprule
Scenario & Construction & Expected structural effect \\
\midrule
Smooth forecast & 
    Gaussian smoothing ($\sigma = 15$ samples) & 
    R-peaks flattened; dominant cycle destroyed; 
    $S_{L_{\max}}$ and $S_{TP}$ collapse \\[6pt]
Attenuated peaks & 
    Amplitudes scaled to 25\% around signal mean & 
    Cycle structure preserved spatially; $S_{L_{\max}}$ 
    and $S_{TP}$ collapse proportionally \\[6pt]
Missing beat & 
    Beat 2 replaced by linear interpolation & 
    One cyclic structure removed; $\beta_1$ drops 
    and temporal overlap degrades \\[6pt]
Phase shift & 
    Signal cyclically shifted by half a beat period & 
    Cyclic structure fully intact; TFS 
    remains high; LTFS collapses \\
\botrule
\end{tabular}
\end{table}

\begin{table}[!b]
\caption{Synthetic ECG results.}
\label{tab:ecg_results}
\small
\setlength{\tabcolsep}{4pt}
\begin{tabular}{@{}lcccccccc@{}}
\toprule
Scenario & MSE $\downarrow$ & 
$S_{\beta_1}$ $\uparrow$ & $S_{L_{\max}}$ $\uparrow$ & 
$S_{TP}$ $\uparrow$ & $S_H$ $\uparrow$ & 
TFS $\uparrow$ & Overlap $\uparrow$ & 
LTFS $\uparrow$\\
\midrule
GT baseline      & 0.0000 & 1.0000 & 1.0000 & 1.0000 
                 & 1.0000 & 1.0000 & 1.0000 & 1.0000 \\
Smooth forecast  & 0.0110 & 0.2780 & 0.0357 
                 & 0.0287 & 0.6627 & 0.1172 
                 & 0.0000 & 0.0000 \\
Attenuated peaks & 0.0081 & 0.9918 & 0.2500 
                 & 0.2500 & 0.9999 & 0.4990 
                 & 1.0000 & 0.4990 \\
Missing beat     & 0.0020 & 0.8622 & 1.0000 
                 & 0.9546 & 0.9342 & 0.9364 
                 & 0.8806 & 0.8246 \\
Phase shift      & 0.0321 & 1.0000 & 1.0000 
                 & 0.9981 & 0.9978 & 0.9990 
                 & 0.0000 & 0.0000 \\
\botrule
\end{tabular}
\end{table}

Table~\ref{tab:ecg_results} reports MSE, TFS, 
LTFS, dominant cycle overlap, and the four 
component scores for each scenario. The GT baseline row confirms pipeline validity; all forecast rows should be read against it.

\subsection{Interpretation}
\label{sec:synthetic_interp}
\textbf{MSE and structural fidelity measure distinct properties.} The smooth forecast achieves MSE $= 0.011$ yet $TFS = 0.117$, 
indicating near-total structural collapse: R-peak topology is 
destroyed while MSE reports a numerically close forecast. 
Conversely, the phase shift scenario achieves $TFS = 0.999$ with MSE $= 0.032$, confirming that the two metrics measure fundamentally different properties.

\textbf{$LTFS$ detects temporal mislocalisation 
invisible to $TFS$.} The phase shift scenario achieves 
$TFS= 0.999$ but $LTFS = 0.000$ with 
Overlap $= 0.000$: the dominant cycle exists at the correct 
amplitude and frequency but at entirely wrong time steps, a 
failure mode invisible to all diagram-level metrics.

\textbf{Component scores decompose structural damage precisely.} 
For attenuated peaks, $S_{L_{\max}} = 0.250$ and $S_{TP} = 0.250$ 
precisely reflect the 25\% amplitude scaling while Overlap $= 
1.000$ confirms correct temporal localisation. For missing beat, 
MSE $= 0.002$ is essentially zero yet $LTFS = 0.829$ 
and Overlap $= 0.881$ reveal the structural gap, invisible to 
pointwise evaluation.

\section{Results}
\label{sec:results}
\subsection{Structural Fidelity Across Datasets and Horizons}
Tables~\ref{tab:scores_48}--\ref{tab:scores_longest} report 
$LTFS$ and its components alongside MSE and MAE 
across all three datasets and prediction horizons. Across all 
datasets, $LTFS$ values remain substantially below 
1.0 even for the best-performing models, indicating that no 
architecture fully preserves both diagram-level structural 
properties and temporal localisation of the dominant cycle 
simultaneously. The overlap component drives $LTFS$ 
lower on Exchange and ILI, confirming that temporal 
mislocalisation is a pervasive failure mode independent of 
model class.

\begin{table*}[!t]
\small
\caption{Forecast $H_1$ descriptors $\beta_1$ and $L_{\max}$ across datasets and prediction horizons. GT rows show ground-truth values.}
\label{tab:descriptors_b1_lmax}
\centering
\setlength{\tabcolsep}{3pt}
\begin{tabular}{llccccccccc}
\toprule
&
&
\multicolumn{3}{c}{ETTm2} &
\multicolumn{3}{c}{Exchange} &
\multicolumn{3}{c}{ILI} \\
\cmidrule(lr){3-5}
\cmidrule(lr){6-8}
\cmidrule(lr){9-11}
Model & Metric
& 48 & 96 & 192
& 48 & 96 & 192
& 36 & 48 & 60 \\
\midrule

\multirow{2}{*}{GT}
& $\beta_1$
& 11.20 & 30.28 & 81.67
& 4.71 & 12.34 & 27.23
& 1.98 & 3.72 & 5.36 \\
& $L_{\max}$
& 0.81 & 1.25 & 1.40
& 0.65 & 0.83 & 0.79
& 0.86 & 1.49 & 2.15 \\
\midrule

\multirow{2}{*}{Autoformer}
& $\beta_1$
& 8.50 & 21.93 & 39.49
& 10.56 & 19.51 & 43.45
& 2.96 & 4.16 & 6.40 \\
& $L_{\max}$
& 1.33 & 1.61 & 2.51
& 1.79 & 1.38 & 1.26
& 1.13 & 1.64 & 1.57 \\
\midrule

\multirow{2}{*}{FEDformer}
& $\beta_1$
& 4.34 & 13.93 & 4.34
& 10.09 & 25.55 & 96.04
& 1.50 & 3.42 & 5.46 \\
& $L_{\max}$
& 1.16 & 2.08 & 1.16
& 0.75 & 0.63 & 0.57
& 0.35 & 0.73 & 1.12 \\
\midrule

\multirow{2}{*}{Transformer}
& $\beta_1$
& 1.16 & 6.01 & 18.27
& 2.68 & 20.65 & 41.89
& 1.06 & 3.44 & 4.94 \\
& $L_{\max}$
& 0.11 & 1.35 & 2.15
& 0.24 & 0.77 & 0.60
& 0.14 & 0.24 & 0.37 \\
\midrule

\multirow{2}{*}{Informer}
& $\beta_1$
& 8.22 & 18.85 & 25.47
& 8.80 & 13.88 & 27.23
& 5.58 & 9.06 & 12.46 \\
& $L_{\max}$
& 0.60 & 0.96 & 0.69
& 0.73 & 0.75 & 0.79
& 0.72 & 0.67 & 0.71 \\
\midrule

\multirow{2}{*}{PatchTST}
& $\beta_1$
& 4.43 & 16.69 & 56.07
& 18.50 & 56.14 & 136.16
& 2.58 & 4.74 & 8.50 \\
& $L_{\max}$
& 0.76 & 2.11 & 2.54
& 0.65 & 0.74 & 0.76
& 0.28 & 0.60 & 1.31 \\
\bottomrule
\end{tabular}
\end{table*}

\begin{table*}[!t]
\small
\caption{Forecast descriptors $TP$ and $H$ across datasets and prediction horizons.}
\label{tab:descriptors_tp_h}
\centering
\setlength{\tabcolsep}{3pt}
\begin{tabular}{llccccccccc}
\toprule
&
&
\multicolumn{3}{c}{ETTm2} &
\multicolumn{3}{c}{Exchange} &
\multicolumn{3}{c}{ILI} \\
\cmidrule(lr){3-5}
\cmidrule(lr){6-8}
\cmidrule(lr){9-11}
Model & Metric
& 48 & 96 & 192
& 48 & 96 & 192
& 36 & 48 & 60 \\
\midrule

\multirow{2}{*}{GT}
& $TP$
& 3.14 & 7.95 & 18.93
& 1.35 & 2.93 & 4.72
& 1.02 & 1.88 & 2.97 \\
& $H$
& 1.97 & 2.90 & 3.92
& 0.99 & 1.84 & 2.64
& 0.34 & 0.63 & 0.88 \\
\midrule

\multirow{2}{*}{Autoformer}
& $TP$
& 3.18 & 6.21 & 11.64
& 5.00 & 5.84 & 8.78
& 1.65 & 2.51 & 3.22 \\
& $H$
& 1.44 & 2.40 & 2.77
& 1.80 & 2.33 & 3.09
& 0.55 & 0.77 & 1.23 \\
\midrule

\multirow{2}{*}{FEDformer}
& $TP$
& 1.73 & 5.00 & 1.73
& 2.33 & 4.45 & 13.50
& 0.41 & 1.12 & 2.25 \\
& $H$
& 0.82 & 1.79 & 0.82
& 1.77 & 2.82 & 4.23
& 0.29 & 0.75 & 1.19 \\
\midrule

\multirow{2}{*}{Transformer}
& $TP$
& 0.16 & 1.86 & 5.30
& 0.36 & 3.07 & 4.84
& 0.19 & 0.46 & 0.79 \\
& $H$
& 0.25 & 0.82 & 1.84
& 0.59 & 2.38 & 3.16
& 0.22 & 0.75 & 1.03 \\
\midrule

\multirow{2}{*}{Informer}
& $TP$
& 2.00 & 4.55 & 4.21
& 2.50 & 3.45 & 4.71
& 1.71 & 2.08 & 2.93 \\
& $H$
& 1.60 & 2.40 & 2.74
& 1.77 & 2.18 & 2.63
& 1.22 & 1.75 & 2.10 \\
\midrule

\multirow{2}{*}{PatchTST}
& $TP$
& 1.39 & 5.68 & 14.61
& 4.36 & 12.35 & 25.41
& 0.44 & 1.10 & 2.62 \\
& $H$
& 0.89 & 1.88 & 3.30
& 2.59 & 3.70 & 4.57
& 0.56 & 1.03 & 1.41 \\
\bottomrule
\end{tabular}
\end{table*}

\subsection{Divergence Between Pointwise Accuracy and 
Structural Fidelity}
\label{sec:mse_vs_topology}
The most consequential finding is that MSE and $LTFS$ 
rankings diverge substantially and systematically. On ETTm2 at 
$p=48$, the Transformer achieves MSE competitive with FEDformer 
and Informer yet its $S_{\text{LTFS}}$ is the lowest among all 
models, driven by near-total collapse in $S_{L_{\max}}$ and 
$S_{TP}$: numerically close forecasts while the dominant cyclic 
structure is destroyed, a failure mode entirely invisible to 
pointwise evaluation.

The Exchange dataset provides the sharpest illustration. 
PatchTST achieves the lowest MSE across all three prediction 
horizons yet its $LTFS$ is consistently the lowest 
among all models, driven by severe loop injection 
($\Delta\beta_1 > 0$, Table~\ref{tab:deltas_b1_lmax}) and 
systematically low temporal overlap. The model ranking first 
under pointwise evaluation ranks last under topological 
evaluation — an inverted ordering reflecting a genuine structural 
property: PatchTST generates spurious cyclic structure on a 
dataset whose ground truth has minimal periodicity, a failure mode that MSE neither penalises nor detects.

\begin{table*}[!t]
\footnotesize
\caption{Signed topological descriptor deltas ($\Delta=\text{FC}-\text{GT}$) across datasets and prediction horizons. Negative values indicate structural loss, while positive values indicate over-generation.}
\label{tab:deltas_b1_lmax}
\centering
\setlength{\tabcolsep}{3pt}
\begin{tabular}{llccccccccc}
\toprule
&
&
\multicolumn{3}{c}{ETTm2} &
\multicolumn{3}{c}{Exchange} &
\multicolumn{3}{c}{ILI} \\
\cmidrule(lr){3-5}
\cmidrule(lr){6-8}
\cmidrule(lr){9-11}
Model & Metric
& 48 & 96 & 192
& 48 & 96 & 192
& 36 & 48 & 60 \\
\midrule

\multirow{2}{*}{Autoformer}
& $\Delta\beta_1$
& $-$2.70 & $-$8.35 & $-$42.18
& $+$5.85 & $+$7.17 & $+$16.22
& $+$0.98 & $+$0.44 & $+$1.04 \\
& $\Delta L_{\max}$
& $+$0.52 & $+$0.36 & $+$1.11
& $+$1.15 & $+$0.54 & $+$0.46
& $+$0.27 & $+$0.15 & $-$0.59 \\
\midrule

\multirow{2}{*}{FEDformer}
& $\Delta\beta_1$
& $-$6.86 & $-$16.35 & $-$6.86
& $+$5.38 & $+$13.21 & $+$68.81
& $-$0.48 & $-$0.30 & $+$0.10 \\
& $\Delta L_{\max}$
& $+$0.35 & $+$0.83 & $+$0.35
& $+$0.10 & $-$0.20 & $-$0.23
& $-$0.51 & $-$0.76 & $-$1.03 \\
\midrule

\multirow{2}{*}{Transformer}
& $\Delta\beta_1$
& $-$10.04 & $-$24.27 & $-$63.40
& $-$2.03 & $+$8.31 & $+$14.66
& $-$0.92 & $-$0.28 & $-$0.42 \\
& $\Delta L_{\max}$
& $-$0.70 & $+$0.10 & $+$0.75
& $-$0.41 & $-$0.07 & $-$0.19
& $-$0.72 & $-$1.25 & $-$1.79 \\
\midrule

\multirow{2}{*}{Informer}
& $\Delta\beta_1$
& $-$2.98 & $-$11.43 & $-$56.20
& $+$4.09 & $+$1.54 & $-$12.63
& $+$3.60 & $+$5.34 & $+$7.10 \\
& $\Delta L_{\max}$
& $-$0.22 & $-$0.29 & $-$0.71
& $+$0.08 & $-$0.09 & $-$0.26
& $-$0.14 & $-$0.82 & $-$1.44 \\
\midrule

\multirow{2}{*}{PatchTST}
& $\Delta\beta_1$
& $-$6.77 & $-$13.59 & $-$25.60
& $+$13.79 & $+$43.80 & $+$108.93
& $+$0.60 & $+$1.02 & $+$3.14 \\
& $\Delta L_{\max}$
& $-$0.05 & $+$0.86 & $+$1.14
& $+$0.00 & $-$0.09 & $-$0.03
& $-$0.59 & $-$0.89 & $-$0.84 \\
\bottomrule
\end{tabular}
\end{table*}

\begin{table*}[!ht]
\small
\caption{Signed $H_1$ descriptor deltas ($\Delta =
\text{FC} - \text{GT}$) across datasets and prediction horizons.Negative values indicate structural loss; positive values indicate over-generation.}
\label{tab:deltas_tp_h}
\centering
\setlength{\tabcolsep}{3pt}
\begin{tabular}{llccccccccc}

\toprule
&
&
\multicolumn{3}{c}{ETTm2} &
\multicolumn{3}{c}{Exchange} &
\multicolumn{3}{c}{ILI} \\
\cmidrule(lr){3-5}
\cmidrule(lr){6-8}
\cmidrule(lr){9-11}
Model & Metric
& 48 & 96 & 192
& 48 & 96 & 192
& 36 & 48 & 60 \\
\midrule

\multirow{2}{*}{Autoformer}
& $\Delta TP$
& $+$0.04 & $-$1.74 & $-$7.29
& $+$3.65 & $+$2.90 & $+$4.07
& $+$0.63 & $+$0.63 & $+$0.25 \\
& $\Delta H$
& $-$0.52 & $-$0.50 & $-$1.15
& $+$0.81 & $+$0.49 & $+$0.44
& $+$0.21 & $+$0.14 & $+$0.34 \\
\midrule

\multirow{2}{*}{FEDformer}
& $\Delta TP$
& $-$1.41 & $-$2.95 & $-$1.41
& $+$0.97 & $+$1.52 & $+$8.79
& $-$0.61 & $-$0.76 & $-$0.71 \\
& $\Delta H$
& $-$1.15 & $-$1.11 & $-$1.15
& $+$0.78 & $+$0.98 & $+$1.59
& $-$0.05 & $+$0.11 & $+$0.31 \\
\midrule

\multirow{2}{*}{Transformer}
& $\Delta TP$
& $-$2.98 & $-$6.09 & $-$13.63
& $-$0.99 & $+$0.13 & $+$0.12
& $-$0.83 & $-$1.41 & $-$2.18 \\
& $\Delta H$
& $-$1.72 & $-$2.07 & $-$2.08
& $-$0.40 & $+$0.54 & $+$0.52
& $-$0.13 & $+$0.11 & $+$0.15 \\
\midrule

\multirow{2}{*}{Informer}
& $\Delta TP$
& $-$1.14 & $-$3.40 & $-$14.73
& $+$1.15 & $+$0.51 & $-$2.37
& $+$0.69 & $+$0.20 & $-$0.04 \\
& $\Delta H$
& $-$0.37 & $-$0.50 & $-$1.17
& $+$0.78 & $+$0.34 & $-$0.43
& $+$0.88 & $+$1.12 & $+$1.22 \\
\midrule

\multirow{2}{*}{PatchTST}
& $\Delta TP$
& $-$1.75 & $-$2.28 & $-$4.33
& $+$3.01 & $+$9.41 & $+$20.70
& $-$0.58 & $-$0.78 & $-$0.35 \\
& $\Delta H$
& $-$1.07 & $-$1.01 & $-$0.62
& $+$1.60 & $+$1.86 & $+$1.93
& $+$0.22 & $+$0.40 & $+$0.53 \\
\bottomrule
\end{tabular}
\end{table*}

\subsection{Dataset-Dependent Structural Failure Signatures}
\label{sec:failure_signatures}
Tables~\ref{tab:deltas_b1_lmax} and~\ref{tab:deltas_tp_h} 
reveal that architectural inductive biases produce characteristic dataset-dependent patterns of structural degradation. On ETTm2, 
$\Delta\beta_1 < 0$ for all models across all horizons, indicating systematic loop loss that grows with prediction 
horizon, most severely for the Transformer and Informer whose 
standard attention provides no structural bias to preserve 
periodicity.

On Exchange, the pattern inverts: Autoformer, FEDformer, and 
PatchTST show $\Delta\beta_1 > 0$ across all horizons, 
indicating loop injection on a dataset whose ground-truth 
topology is sparse. The decomposition-based architectures impose 
periodic structure regardless of whether the input supports it. 
PatchTST's injection grows with horizon, reaching $\Delta\beta_1 
= +108.93$ at $p=192$, the largest over-generation observed.

On ILI, failure signatures are more heterogeneous. Informer 
over-generates loops while losing dominant cycle strength — 
additional but weaker cyclic structure rather than the dominant 
one preserved. PatchTST shows a similar pattern; the Transformer 
shows the opposite, with loop count and cycle strength both 
declining progressively, pointing to gradual structural collapse. 
FEDformer is the most consistent, with near-zero $\Delta\beta_1$ 
at $p=60$.

\subsection{Dominant Cycle Overlap and Temporal Phase Errors}
\label{sec:overlap_results}
The overlap component of $LTFS$ reveals a failure 
mode orthogonal to diagram-level structural quality. On ETTm2, 
models show systematic overlap degradation at longer horizons 
even when aggregate structural properties are partially 
preserved. Autoformer maintains the highest overlap across all 
horizons, consistent with its autocorrelation mechanism 
targeting periodic alignment.

On Exchange, overlap is uniformly low across all models and 
horizons. Temporal mislocalisation is the primary failure mode 
on non-stationary data independent of architecture: 
$TFS$ values are moderate for several models yet the 
dominant cycle is placed at systematically wrong time steps, a 
failure visible only through $LTFS$.

On ILI, overlap is more variable. Autoformer achieves the 
highest overlap at $p=36$ (0.60), while Informer and PatchTST 
show lower and declining overlap with horizon length, confirming 
that temporal mislocalisation contributes to structural 
degradation beyond what diagram-level scores alone capture.

\begin{table}[!b]
\caption{Structural quality scores and pointwise errors 
at the shortest prediction horizon per dataset 
(ETTm2 $p=48$, Exchange $p=48$, ILI $p=36$).}
\label{tab:scores_48}
\setlength{\tabcolsep}{3pt}
\begin{tabular}{@{}llrrrrrrrrr@{}}
\toprule
Dataset & Model & 
$S_{\beta_1}$ & 
$S_{L_{\max}}$ & 
$S_{TP}$ & 
$S_{H}$ & 
$TFS$ & 
Overlap & 
$LTFS$ $\uparrow$ & 
MSE $\downarrow$ &
MAE $\downarrow$ \\
\midrule
\multirow{5}{*}{ETTm2}
& Autoformer  & 0.76 & 0.61 & 0.99 & 0.73 & 0.76 & 0.56 & 0.42 & 0.12 & 0.24 \\
& FEDformer   & 0.39 & 0.70 & 0.55 & 0.41 & 0.50 & 0.41 & 0.21 & 0.12 & 0.24 \\
& Transformer & 0.10 & 0.14 & 0.05 & 0.12 & 0.10 & 0.11 & 0.01 & 0.12 & 0.25 \\
& Informer    & 0.73 & 0.73 & 0.64 & 0.81 & 0.73 & 0.26 & 0.19 & 0.18 & 0.31 \\
& PatchTST    & 0.40 & 0.94 & 0.44 & 0.45 & 0.52 & 0.44 & 0.23 & 0.10 & 0.21 \\
\midrule
\multirow{5}{*}{Exchange}
& Autoformer  & 0.45 & 0.36 & 0.27 & 0.55 & 0.39 & 0.55 & 0.22 & 0.11 & 0.24 \\
& FEDformer   & 0.47 & 0.86 & 0.58 & 0.56 & 0.60 & 0.40 & 0.24 & 0.09 & 0.21 \\
& Transformer & 0.57 & 0.37 & 0.27 & 0.59 & 0.43 & 0.23 & 0.10 & 0.40 & 0.49 \\
& Informer    & 0.54 & 0.89 & 0.54 & 0.56 & 0.62 & 0.25 & 0.16 & 0.58 & 0.61 \\
& PatchTST    & 0.25 & 0.99 & 0.31 & 0.38 & 0.42 & 0.27 & 0.11 & 0.04 & 0.14 \\
\midrule
\multirow{5}{*}{ILI}
& Autoformer  & 0.67 & 0.76 & 0.62 & 0.62 & 0.66 & 0.60 & 0.40 & 3.14 & 1.25 \\
& FEDformer   & 0.76 & 0.41 & 0.40 & 0.85 & 0.57 & 0.27 & 0.15 & 2.74 & 1.13 \\
& Transformer & 0.54 & 0.17 & 0.18 & 0.63 & 0.32 & 0.19 & 0.06 & 4.40 & 1.41 \\
& Informer    & 0.35 & 0.84 & 0.60 & 0.28 & 0.47 & 0.39 & 0.18 & 3.78 & 1.38 \\
& PatchTST    & 0.77 & 0.32 & 0.44 & 0.61 & 0.51 & 0.24 & 0.12 & 2.22 & 0.94 \\
\botrule
\end{tabular}
\end{table}

\begin{table}[!t]
\caption{Structural quality scores and pointwise errors 
at $p=96$ (ETTm2, Exchange) and $p=48$ (ILI).}
\label{tab:scores_96}
\setlength{\tabcolsep}{3pt}
\begin{tabular}{@{}llrrrrrrrrr@{}}
\toprule
Dataset & Model & 
$S_{\beta_1}$ & 
$S_{L_{\max}}$ & 
$S_{TP}$ & 
$S_{H}$ & 
$TFS$ & 
Overlap & 
$LTFS$ $\uparrow$ & 
MSE $\downarrow$ &
MAE $\downarrow$ \\
\midrule
\multirow{5}{*}{ETTm2}
& Autoformer  & 0.72 & 0.77 & 0.78 & 0.83 & 0.78 & 0.48 & 0.37 & 0.14 & 0.26 \\
& FEDformer   & 0.46 & 0.60 & 0.63 & 0.62 & 0.57 & 0.45 & 0.26 & 0.13 & 0.25 \\
& Transformer & 0.20 & 0.93 & 0.23 & 0.28 & 0.33 & 0.35 & 0.12 & 0.16 & 0.28 \\
& Informer    & 0.62 & 0.77 & 0.57 & 0.83 & 0.69 & 0.21 & 0.14 & 0.17 & 0.30 \\
& PatchTST    & 0.55 & 0.59 & 0.71 & 0.65 & 0.62 & 0.46 & 0.29 & 0.11 & 0.22 \\
\midrule
\multirow{5}{*}{Exchange}
& Autoformer  & 0.63 & 0.61 & 0.50 & 0.79 & 0.62 & 0.42 & 0.26 & 0.15 & 0.29 \\
& FEDformer   & 0.48 & 0.76 & 0.66 & 0.65 & 0.63 & 0.19 & 0.12 & 0.14 & 0.27 \\
& Transformer & 0.60 & 0.92 & 0.96 & 0.77 & 0.80 & 0.25 & 0.20 & 0.67 & 0.64 \\
& Informer    & 0.89 & 0.90 & 0.85 & 0.85 & 0.87 & 0.14 & 0.12 & 0.77 & 0.70 \\
& PatchTST    & 0.22 & 0.89 & 0.24 & 0.50 & 0.39 & 0.26 & 0.10 & 0.08 & 0.20 \\
\midrule
\multirow{5}{*}{ILI}
& Autoformer  & 0.89 & 0.91 & 0.75 & 0.82 & 0.84 & 0.44 & 0.37 & 3.08 & 1.22 \\
& FEDformer   & 0.92 & 0.49 & 0.60 & 0.85 & 0.69 & 0.33 & 0.23 & 2.69 & 1.12 \\
& Transformer & 0.92 & 0.16 & 0.25 & 0.85 & 0.42 & 0.23 & 0.10 & 4.89 & 1.49 \\
& Informer    & 0.41 & 0.45 & 0.90 & 0.36 & 0.50 & 0.26 & 0.13 & 5.92 & 1.74 \\
& PatchTST    & 0.78 & 0.40 & 0.58 & 0.61 & 0.58 & 0.39 & 0.23 & 2.22 & 0.94 \\
\botrule
\end{tabular}
\end{table}

\begin{table}[!ht]
\caption{Structural quality scores and pointwise errors 
at $p=192$ (ETTm2, Exchange) and $p=60$ (ILI).}
\label{tab:scores_longest}
\setlength{\tabcolsep}{3pt}
\begin{tabular}{@{}llrrrrrrrrr@{}}
\toprule
Dataset & Model & 
$S_{\beta_1}$ & 
$S_{L_{\max}}$ & 
$S_{TP}$ & 
$S_{H}$ & 
$TFS$ & 
Overlap & 
$LTFS$ $\uparrow$ & 
MSE $\downarrow$ &
MAE $\downarrow$ \\
\midrule
\multirow{5}{*}{ETTm2}
& Autoformer  & 0.48 & 0.56 & 0.61 & 0.71 & 0.58 & 0.46 & 0.27 & 0.15 & 0.28 \\
& FEDformer   & 0.39 & 0.70 & 0.55 & 0.41 & 0.50 & 0.41 & 0.21 & 0.12 & 0.24 \\
& Transformer & 0.22 & 0.65 & 0.28 & 0.47 & 0.37 & 0.38 & 0.14 & 0.22 & 0.32 \\
& Informer    & 0.31 & 0.49 & 0.22 & 0.70 & 0.39 & 0.13 & 0.05 & 0.97 & 0.77 \\
& PatchTST    & 0.69 & 0.55 & 0.77 & 0.84 & 0.70 & 0.50 & 0.35 & 0.15 & 0.26 \\
\midrule
\multirow{5}{*}{Exchange}
& Autoformer  & 0.63 & 0.63 & 0.54 & 0.86 & 0.65 & 0.26 & 0.17 & 0.34 & 0.42 \\
& FEDformer   & 0.28 & 0.71 & 0.35 & 0.62 & 0.46 & 0.10 & 0.04 & 0.26 & 0.37 \\
& Transformer & 0.65 & 0.76 & 0.98 & 0.84 & 0.80 & 0.12 & 0.09 & 1.12 & 0.82 \\
& Informer    & 0.54 & 0.66 & 0.50 & 0.83 & 0.62 & 0.09 & 0.05 & 2.52 & 1.32 \\
& PatchTST    & 0.20 & 0.96 & 0.19 & 0.58 & 0.38 & 0.13 & 0.05 & 0.18 & 0.30 \\
\midrule
\multirow{5}{*}{ILI}
& Autoformer  & 0.84 & 0.73 & 0.92 & 0.72 & 0.80 & 0.44 & 0.35 & 3.08 & 1.22 \\
& FEDformer   & 0.98 & 0.52 & 0.76 & 0.74 & 0.73 & 0.31 & 0.23 & 2.71 & 1.12 \\
& Transformer & 0.92 & 0.17 & 0.27 & 0.85 & 0.44 & 0.27 & 0.12 & 5.34 & 1.55 \\
& Informer    & 0.43 & 0.33 & 0.99 & 0.42 & 0.49 & 0.19 & 0.09 & 7.25 & 1.94 \\
& PatchTST    & 0.63 & 0.61 & 0.88 & 0.62 & 0.68 & 0.28 & 0.19 & 2.18 & 0.95 \\
\botrule
\end{tabular}
\end{table}


\subsection{Comprehensive Metric Comparison}
\label{sec:metric_comparison}

Tables~\ref{tab:full_metrics_ili_60} 
and~\ref{tab:full_metrics_exchange_192} report pointwise 
accuracy (MSE, RMSE, MAE, MASE, RMSSE), shape similarity 
(DTW), temporal alignment (TDI), distribution distance 
($W_2$), and topological fidelity ($LTFS$, 
Overlap) for ILI ($p=60$) and Exchange ($p=192$).

On ILI, PatchTST ranks first under all conventional metrics 
yet its $LTFS$ (0.19) ranks fourth. Autoformer 
ranks third under MSE but achieves the highest 
$LTFS$ (0.35) and Overlap (0.44). On Exchange, 
the divergence is more pronounced: PatchTST again leads all 
pointwise metrics yet achieves the highest $W_2$ (12.633) 
and joint lowest $LTFS$ (0.05), consistent 
with the severe loop injection identified in 
Section~\ref{sec:failure_signatures}. FEDformer shows the 
highest TDI (52.145) on Exchange despite competitive MSE, 
indicating large temporal displacements invisible to 
pointwise and shape-similarity metrics.

\textbf{Agreement with existing structural metrics.} 
$LTFS$ shows broad directional agreement with 
$W_2$ and DTW on model rankings. On ILI, models with lower 
DTW (PatchTST: 8.347, FEDformer: 8.701) tend to achieve 
higher $LTFS$ relative to models with higher 
DTW (Informer: 18.376, Transformer: 14.959), confirming 
that topological fidelity and shape similarity are 
correlated dimensions of structural quality. Similarly, 
lower $W_2$ values broadly correspond to higher 
$LTFS$ on ILI, with Autoformer and FEDformer 
occupying the top positions on both metrics. This agreement 
is consistent with prior work showing that persistence 
diagram distances and shape-based metrics capture 
overlapping but non-identical aspects of signal 
structure~\cite{perea2015sliding}.

\textbf{Enhancement over existing metrics.} Despite this 
directional agreement, $LTFS$ provides 
additional discriminative information not available from 
DTW, TDI, or $W_2$ alone. On Exchange, PatchTST achieves 
the lowest DTW (4.477) yet the highest $W_2$ (12.633) and 
lowest $S_{\text{LTFS}}$ (0.05), a pattern that neither 
DTW nor $W_2$ alone can expose. The Overlap component of 
$LTFS$ further distinguishes models that 
preserve diagram-level structure from those that additionally 
preserve the temporal location of the dominant oscillation: 
on ILI, Autoformer and FEDformer achieve comparable DTW 
(10.253 vs 8.701) yet substantially different Overlap 
(0.44 vs 0.31), a difference invisible to all scalar 
distance metrics. Furthermore, the four-component 
decomposition of $TFS$ allows the source of 
structural degradation to be identified precisely — whether 
it originates from loop count loss, dominant cycle 
attenuation, total persistence reduction, or diagram 
complexity change — a diagnostic capability absent from 
DTW, TDI, and $W_2$. Together, these properties establish 
$LTFS$ as a complementary evaluation signal 
that both corroborates and extends the information provided 
by existing structural metrics.

\begin{table}[!t]
\caption{Comprehensive evaluation metrics for ILI 
at $p=60$.}
\label{tab:full_metrics_ili_60}
\setlength{\tabcolsep}{3pt}
\footnotesize
\begin{tabular}{@{}lrrrrrrrrrr@{}}
\toprule
Model & 
MSE $\downarrow$ & 
RMSE $\downarrow$ & 
MAE $\downarrow$ & 
MASE $\downarrow$ & 
RMSSE $\downarrow$ & 
DTW $\downarrow$ & 
TDI $\downarrow$ & 
$W_2$ $\downarrow$ & 
Overlap $\uparrow$ & 
$S_{\text{LTFS}}$ $\uparrow$ \\
\midrule
Autoformer  &  3.027 & 1.679 & 1.178 & 10.244 & 11.659 & 10.253 & 6.466 & 3.621 & 0.44 & 0.35 \\
FEDformer   &  2.712 & 1.589 & 1.121 &  9.681 & 10.934 &  8.701 & 6.016 & 3.410 & 0.31 & 0.23 \\
Transformer &  5.342 & 2.244 & 1.553 & 11.846 & 14.990 & 14.959 & 9.986 & 4.977 & 0.27 & 0.12 \\
Informer    &  7.223 & 2.627 & 1.949 & 13.046 & 15.840 & 18.376 & 7.494 & 5.491 & 0.19 & 0.09 \\
PatchTST    &  2.178 & 1.413 & 0.946 &  7.699 &  9.557 &  8.347 & 4.300 & 4.706 & 0.28 & 0.19 \\
\botrule
\end{tabular}
\end{table}

\begin{table}[!b]
\caption{Comprehensive evaluation metrics for Exchange 
at $p=192$.}
\label{tab:full_metrics_exchange_192}
\setlength{\tabcolsep}{3pt}
\footnotesize
\begin{tabular}{@{}lrrrrrrrrrr@{}}
\toprule
Model & 
MSE $\downarrow$ & 
RMSE $\downarrow$ & 
MAE $\downarrow$ & 
MASE $\downarrow$ & 
RMSSE $\downarrow$ & 
DTW $\downarrow$ & 
TDI $\downarrow$ & 
$W_2$ $\downarrow$ & 
Overlap $\uparrow$ & 
$S_{\text{LTFS}}$ $\uparrow$ \\
\midrule
Autoformer  & 0.337 & 0.545 & 0.420 & 20.555 & 16.584 &  5.528 & 47.024 &  6.669 & 0.26 & 0.17 \\
FEDformer   & 0.262 & 0.487 & 0.373 & 17.906 & 14.403 &  5.017 & 52.145 &  5.715 & 0.10 & 0.08 \\
Transformer & 1.123 & 1.032 & 0.820 & 48.939 & 35.863 & 10.019 & 46.922 &  9.740 & 0.12 & 0.09 \\
Informer    & 2.852 & 1.650 & 1.389 & 84.057 & 59.865 & 18.766 & 28.876 &  8.746 & 0.09 & 0.05 \\
PatchTST    & 0.179 & 0.403 & 0.298 & 14.443 & 12.192 &  4.477 & 49.778 & 12.633 & 0.13 & 0.05 \\
\botrule
\end{tabular}
\end{table}

\section{Discussion}
\label{sec:discussion}

The results across all three datasets and five architectures 
point to three consistent observations worth synthesising.

\textbf{Topological and pointwise evaluation are 
complementary, not redundant.} The divergence between MSE 
and $LTFS$ rankings is not an artefact of 
a specific dataset or horizon — it appears consistently 
across ETTm2, Exchange, and ILI, and at all three 
prediction lengths. This suggests that structural fidelity 
and numerical accuracy reflect genuinely different 
properties of forecast quality, and that neither subsumes 
the other. A practitioner relying solely on MSE would 
systematically misrank architectures on structural grounds, 
and vice versa.

\textbf{Architectural inductive biases have topological 
consequences.} The dataset-dependent failure signatures 
identified in Section~\ref{sec:failure_signatures} are 
not random — they are predictable from the inductive bias 
of each architecture. Decomposition-based models 
(Autoformer, FEDformer) inject loops on non-stationary 
data because their seasonal components impose periodic 
structure regardless of the signal. Standard attention 
models (Transformer, Informer) lose loops on periodic 
data because they lack any structural prior to preserve 
cyclicity. PatchTST's patch-based tokenisation amplifies 
loop injection at long horizons. These patterns suggest 
that topological evaluation could inform architecture 
selection for structurally sensitive applications.

\textbf{Temporal localisation is a distinct failure 
dimension.} The gap between $TFS$ and 
$LTFS$ across all experiments confirms that 
preserving diagram-level topological structure is 
insufficient — a forecast can reproduce the correct loop 
count, dominant cycle strength, and total persistence 
while placing the dominant oscillation at entirely wrong 
time steps. This failure mode is orthogonal to both 
pointwise accuracy and existing structural metrics 
including $W_2$ and DTW, and is only made visible through 
the dominant cycle overlap component.

\section{Conclusion}
\label{sec:conclusion}

We presented TopoCast, a persistent homology-based framework 
for evaluating structural fidelity of time series forecasts 
as a complement to pointwise metrics. Takens delay embedding 
lifts forecast and ground-truth sequences into phase space, 
from which $H_1$ persistence diagrams yield four structural 
quality scores combined into $LTFS$ via 
geometric mean and dominant cycle overlap.

Experiments across five Transformer architectures and three 
benchmarks show that topological and pointwise rankings 
diverge substantially, that architectural inductive biases 
produce characteristic structural failure signatures, and 
that dominant cycle overlap detects temporal phase errors 
invisible to all diagram-level metrics. TopoCast requires 
no model access and exposes a dimension of forecast quality 
that MSE cannot measure. Future work includes higher homology 
dimensions, probabilistic forecasts, and topology-aware 
training objectives.

\bibliography{sn-bibliography}

\end{document}